\documentclass{article}

\usepackage{PRIMEarxiv}

\usepackage{multicol}
\usepackage{amsmath}
\usepackage[utf8]{inputenc} % allow utf-8 input
\usepackage[T1]{fontenc}    % use 8-bit T1 fonts
\usepackage{hyperref}       % hyperlinks
\usepackage{url}            % simple URL typesetting
\usepackage{booktabs}       % professional-quality tables
\usepackage{amsfonts}       % blackboard math symbols
\usepackage{nicefrac}       % compact symbols for 1/2, etc.
\usepackage{microtype}      % microtypography
\usepackage{lipsum}
\usepackage{fancyhdr}       % header
\usepackage{graphicx}       % graphics
\graphicspath{{media/}}     
\usepackage{wrapfig}
\usepackage{graphicx}
\usepackage{subfigure}
\usepackage{multirow}
\usepackage{xcolor}
\usepackage{algorithm}
\usepackage{algorithmic}
\usepackage{adjustbox}
\usepackage{multirow}

\title{Improved TokenPose with Sparsity
}

\author{
  Anning Li \\
  University of Electronic Science and Technology of China \\
  \texttt{anning@std.uestc.edu.cn} \\
}

\begin{document}
\maketitle

\begin{abstract}
Over the past few years, the vision transformer and its various forms have gained significance in human pose estimation. By treating image patches as tokens, transformers can capture global relationships wisely, estimate the keypoint tokens by leveraging the visual tokens, and recognize the posture of the human body. Nevertheless, global attention is computationally demanding, which poses a challenge for scaling up transformer-based methods to high-resolution features. In this paper, we introduce sparsity in both keypoint token attention and visual token attention to improve human pose estimation. Experimental results on the MPII dataset demonstrate that our model has a higher level of accuracy and proved the feasibility of the method, achieving new state-of-the-art results. The idea can also provide references for other transformer-based models. 
\end{abstract}

% keywords can be removed
\keywords{Vision Transformer \and Human Pose Estimation \and Machine Learning \and Self-attention mechanism}

\section{Introduction}
Human pose estimation Is the process of recovering anatomical keypoints of the human body given an image or a video, which deeply relies on both visual cues and keypoints constraint relationships. The most common approach for human pose estimation is to use convolutional neural networks (CNNs) to extract features from the input image \cite{toshev2014deeppose,wei2016convolutional,newell2016stacked,xiao2018simple,sun2019deep,bestofboth,wang2020predicting}. However, they have some limitations and challenges. Human body joints are highly correlated, and constrained by strong kinetic and physical restrictions. \cite{tompson2014joint}. Therefore, the limited reception fields of convolutional neural networks (CNNs) can result in poor capture of the long-range constraints that exist between joints\cite{li2021tokenpose}.

The transformer-based Vision Transformer \cite{dosovitskiy2020image} and its variants have showcased impressive performance on various visual tasks\cite{touvron2020training,carion2020end}. In comparison with CNN, the self-attention mechanism of transformers can capture long-range dependencies\cite{Bello2019ICCV} and learn the relationship among keypoints better. In single-view 2D human pose estimation, TransPose \cite{yang2020transpose} and TokenPose \cite{li2021tokenpose} have achieved new state-of-the-art performances. Typically, the feature maps are flattened into unidimensional token sequences and then fed into the transformer. TokenPose divides the keypoints into two categories, visual tokens and keypoint tokens. Visual tokens generally refer to the visual features or representations used to encode the appearance and context information in an image or video frame. Keypoint tokens, on the other hand, refer to the specific body keypoints that are of interest in the pose estimation task, represented as coordinates $(x, y)$ or $(x, y, z)$ in the image or 3D space. 

However, The computation complexity is expensive due to dimensional transformation, with the amount being quadratic to the length of input sequences. Therefore, it is challenging to extend these methods to high-resolution feature maps. What's more, the attention maps are generally very sparse and only attend to a small local region. Most of the redundant areas can be mowed to speed up the training process. This also prompts our interest to strengthen the connections between keypoint tokens, since we can generalize mathematical matrices through certain joint contacts in human anatomy.

\begin{figure*}[h]
    \centering
    \includegraphics[width=1.0\linewidth]{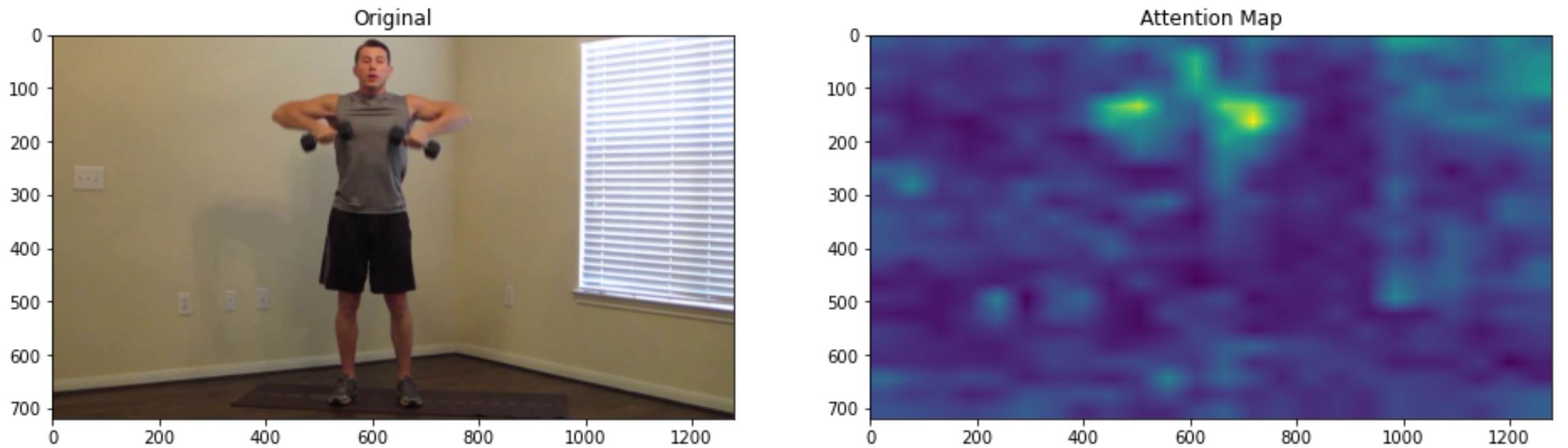}
    \caption{\small{Attention map is usually very sparse and only attends to a small local region.}}
    \label{fig:introduction}
\end{figure*}

In this paper, we propose a more efficient, yet accurate Transformer model based on the TokenPose Transformer\cite{li2021tokenpose}, named Improve TokenPose with Sparsity. Specifically, we focus on two types of sparse tokens, human-joint keypoint tokens, and attention-level visual tokens. On keypoint tokens, we refer to human physiological structures to strengthen the connection of relevant joints with a sparse joint mask. On visual tokens, we introduce sparsity at the attention level by pruning attention connections. We reveal that pruning these sparse tokens does not reduce the accuracy but can accelerate the entire network. 
 
Our main contributions to the Improved TokenPose with Sparsity are summarized as follows:
\begin{enumerate}
\item We propose an attention-based pruning strategy to dynamically upgrade a sparse attention mask. By focusing on the more significant visual tokens, we can simultaneously achieve both token and attention connections during training while making use of their sparsities. Our method does not require a complex token selector module or additional training loss or hyper-parameters.
\item We propose a self-defined sparse joint mask based on human structure. With the use of graph degree, we can pre-define a 0-1 mask to emphasize the connections between adjacent joints and symmetrical joints. 
\item We conduct extensive experiments on MPII dataset, and prove the feasibility of the method by experimental analysis.

\end{enumerate}

\vspace{0.5em}
\section{Related Work}
\subsection{Vision Transformer}
Vision transformer \cite{vaswani2017attention} was originally proposed for natural language processing, which has been shown to be effective in capturing long-range dependencies in sequences of tokens. The key idea of ViT is to apply the Transformer architecture directly to image patches. Recently, they have shown remarkable success in various applications such as classification \cite{dosovitskiy2020image,touvron2020training}, object detection\cite{carion2020end,zhu2020deformable,fang2021you}, and semantic segmentation\cite{zheng2020rethinking,wang2020end,yan2022after,you2022class}. 
Since its introduction, ViT has achieved state-of-the-art performance on several benchmark datasets for image recognition, including ImageNet\cite{Russakovsky2014ImageNetLS}. ViT has also inspired further research in the area of vision transformers, including ViT variants with improved performance, such as DeiT \cite{pmlrv139touvron21a}, and methods for efficient training of vision transformers, such as TNT\cite{dang2022tnt}. However, their extensive computation and memory usage during both training and inference hinder their generalization. Existing compression algorithms typically focus on efficient inference and rely on pre-trained dense models, neglecting time-consuming training. Researchers have proposed several methods to improve the efficiency of ViT. For example,  Network pruning\cite{han2015deep,chen2021chasing,chen2022principle,yu2022unified}, the vanilla ViT \cite{dosovitskiy2020image}, knowledge distillation \cite{hinton2015distilling,touvron2020training,chen2022dearkd}, and quantization\cite{shen2020q,sun2022vaqf} have been applied to ViT, and other algorithms introduce CNN properties to alleviate the burden of computing global attention\cite{liu2021swin,chen2021crossvit}.

Additionally, some models slim the input tokens by aggregating neighboring tokens\cite{yuan2021tokens,caron2021emerging,ryoo2021tokenlearner,rao2021dynamicvit,kong2021spvit,liang2022evit,meng2022adavit}. Their core idea is mining tokens using learnable attention weights or reducing and reorganizing image tokens based on the classification token. For example the Token-to-tokens\cite{yuan2021tokens}, the TokenLearner,\cite{ryoo2021tokenlearner}, the DynamicViT \cite{rao2021dynamicvit} and the EViT \cite{liang2022evit}. These models, however, have only been designed for classification tasks, where the ultimate prediction relies solely on the classification token, whereas in human pose estimation, it is also important to take spatial clues and articular association into consideration. 

\subsection{Human Pose Estimation}
Human pose estimation is an important research area in computer vision that involves estimating the joint locations and orientations of a person's body from images or videos. Previously, many CNN models were proposed in 2D human pose estimation\cite{wei2016convolutional,xiao2018simple,sun2019deep}. Despite its recognized effectiveness in capturing local information, convolutional neural networks still struggle with modeling long-range relationships. This is particularly challenging when it comes to human pose estimation. Therefore, many works have been done to explore efficient architecture design for human pose estimation. Such as HRNet\cite{sun2019deep}, is a multi-stage, high-resolution network designed specifically for human pose estimation. It uses a high-resolution feature extraction stage, followed by a set of parallel branches that combine information across different scales to achieve state-of-the-art results. Lite-HRNet \cite{yu2021lite} is a lightweight version of the HRNet with limited computational resources. It achieves this by reducing the number of channels and layers in the original HRNet while retaining its multi-resolution feature extraction and integration capabilities. Other works are also designed for real-time pose estimation \cite{osokin2018real,neff2020efficienthrnet}. However, their focus is primarily on CNN-based networks, and none of their studies explore transformer-based networks.

Recently, many works have used transformer networks for human pose estimation. For example, TransPose utilizes transformers to illustrate the interdependence of keypoint predictions\cite{yang2020transpose}, TokenPose incorporates extra keypoint tokens to acquire knowledge of constraint relationships and appearance cues\cite{li2021tokenpose} and employs a transformer module to capture the long-range dependencies between body parts in human pose estimation for video object segmentation\cite{yin2020disentangled}. These works highlight the growing use of transformer-based models in human pose estimation. By leveraging self-attention mechanisms and capturing global dependencies, transformer-based approaches have demonstrated promising results and opened up new avenues for advancing the accuracy and robustness of pose estimation systems.

Both CNNs and Transformer models are effective in human pose estimation tasks. CNNs are well-suited for extracting local features and handling spatial variability, while Transformers are better at capturing complex global relationships between body parts. Both emphasize focus on global information, efficiency, and robustness to noise.

\section{Methodology}
\subsection{Overview}

\begin{figure}[ht]
    \centering
    \includegraphics[width=1.0\linewidth]{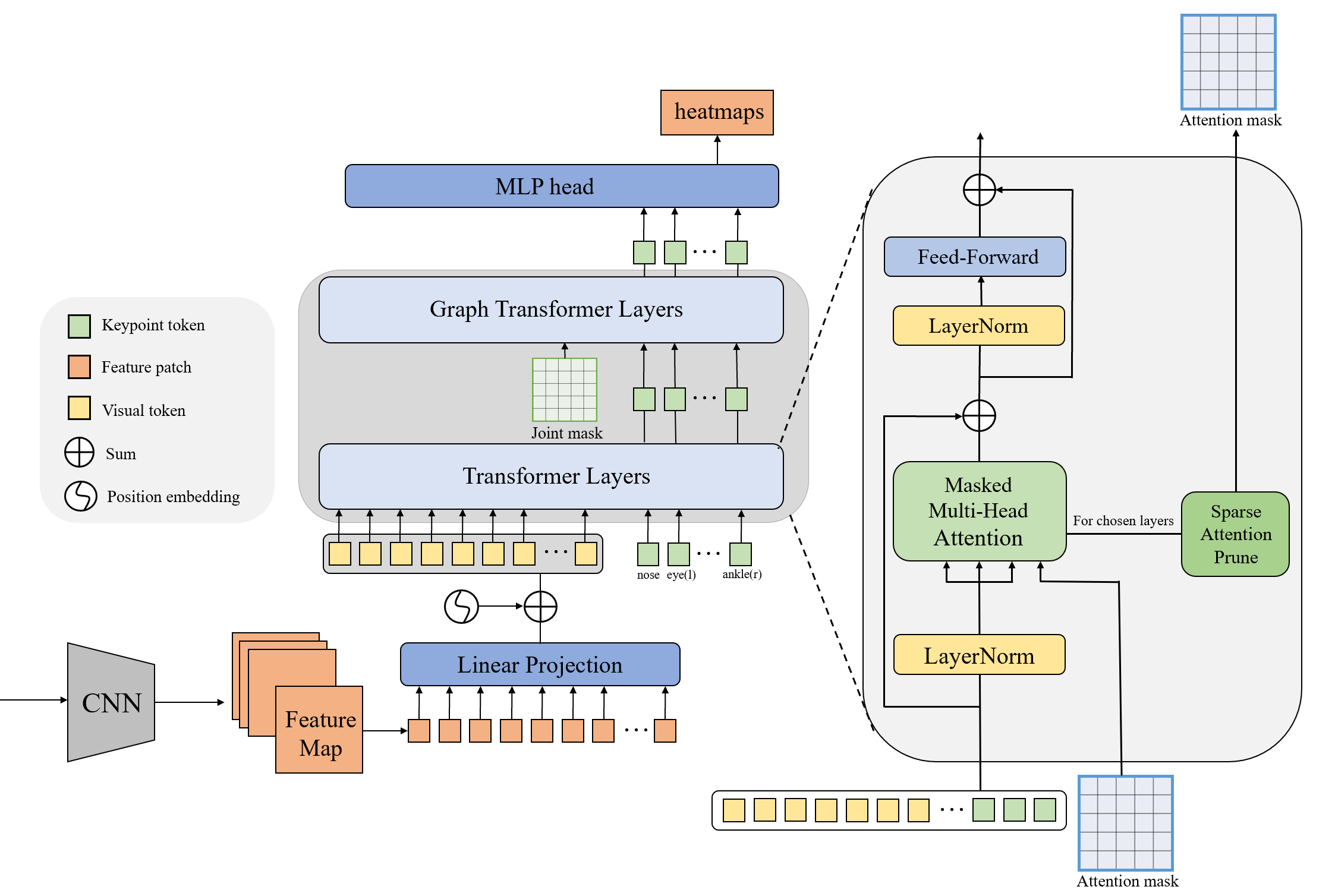}
    \caption{\small{A schematic diagram of the Improved TokenPose.} }
    \label{fig:framework1}
\end{figure}

Fig.\ref{fig:framework1} is an overview of our Improved TokenPose with Sparsity. The input image $\mathbf{I}$ first goes through a CNN backbone and outputs the feature map $\mathbf{F} \in  \mathbb{R}^{ H \times W \times C}$. Then the feature map is split into patches $\mathbf{F}_p \in \mathbb{R}^{N_P \times (C \cdot P_h \cdot P_w) }$ and flattened to 1D vectors $\mathbf{X}_p \in \mathbb{R}^{N_v \times D}$. $(P_h, P_w)$ is the resolution of each image patch. $D$ is the dimension of hidden embeddings. $ N_p = \frac{H}{P_h} \cdot \frac{W}{P_w}$ is the total number of patches. Then, by using linear projection, the flattened vectors produce visual tokens $\mathbf{X}_v = \mathbf{X}_p + \mathbf{E}$. $\mathbf{E}$ is the 2D positional encoding $\mathbf{E} \in \mathbb{R}^{N_p \times D}$. $J$ specific types of tokens are randomly initialized to represent individual keypoint tokens $\mathbf{X}_k \in \mathbb{R}^{J \times D}$. The visual tokens and keypoint tokens are then fed into the Transformer encoder. The input to the transformer is $\mathbf{X}^0 = [\mathbf{X}_k, \mathbf{X}_v] \in  \mathbb{R}^{(N_p+J) \times D}$.

The transformer has $L$ encoder layers in total. The encoder layer consists of multi-headed self-attention. The multi-head attention in each layer captures both appearance and anatomical constraint cues. At chosen $L_3^{th}, L_6^{th}, L_9^{th}$ layers, we employ the masked multi-head self-attention mechanism to make use of the sparsity in visual token attention and better measure the complexity of images(see 3.2).

After the $L$ encoder layers, the output keypoint tokens are then put into the $L_1 (L_1 \textless L)$ layers graph transformer with a self-defined sparse joint mask $\mathbf{M_J} \in  \mathbb{R}^{ J \times J}$, which is initialized to represent the connection of human joints. The graph transformer has the same structure as the transformer, only this time we consider the joint mask as the initialized attention mask. Through the graph transformer layers, we once again emphasize the relevance of the human body structure. In the end, an MLP head uses the keypoint tokens $\mathbf{X}_k^L \in \mathbb{R}^{J\times D}$ from the last Graph Transformer layer to predict the keypoints heatmaps $\mathbf{H} \in \mathbb{R}^{J\times (H_h\cdot W_h)}$.
\vspace{-0.5em}

\begin{figure}[h]
    \centering
    \includegraphics[width=0.4\linewidth]{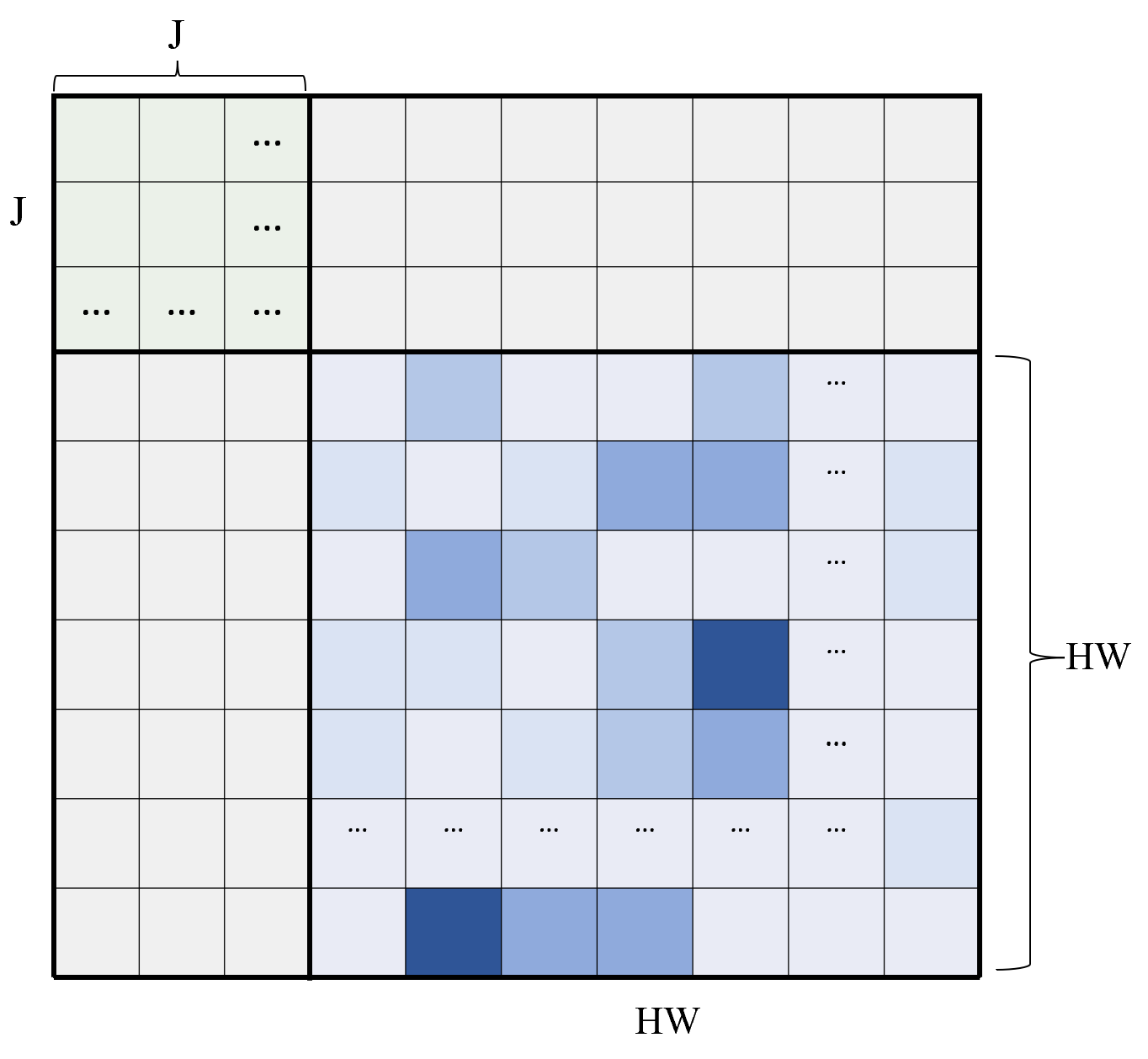}
    \caption{\small{Image map}}
    \label{fig:framework2}
\end{figure}

Fig.\ref{fig:framework2} is a schematic diagram of our proceed tokens. Based on the TokenPose model\cite{li2021tokenpose}, in order to improve the accuracy of its application in human pose estimation, we propose improvements in two places mainly focusing on the upper left keypoint token joint mask and lower right visual token attention mask. Since the image size passed to the transformer encoder is ${(HW+J) \times (HW+J)}$, we can abstract a diagram like Fig.\ref{fig:framework2} to help us understand the tokens we are processing. By pruning the two areas ${HW\times HW}$ and ${J \times J}$, and using the sparsity of the chosen tokens, we can remove the relatively less important data and still get an accurate ultimate result. 

\subsection{Masked Multi-head Self-attention(MMSA)}

The Masked Multi-head Self-attention aims to match a query and a set of key-value pairs to an output\cite{vaswani2017attention}. Given the inputs $\mathbf{X}$ and $\mathbf{M}$, three linear projections are applied to transfer $\mathbf{X}$ into three matrices of equal size, namely the query $\mathbf{Q}$, the key $\mathbf{K}$, and the value $\mathbf{V}$. The Masked Multi-head Self-attention (MMSA) operation is calculated by:

\begin{equation}
    \text{MMSA}(\mathbf{X}) =  \text{Softmax}( \frac{ \mathbf{Q} \mathbf{K}^T \times \mathbf{M}}{\sqrt{D}}  )\mathbf{V},
    \label{eq:attention1}
\end{equation}

According to Fig.\ref{fig:framework1}, the tokens go through two different Transformer encoders in turn. For the first Transformer encoder, in the process of MMSA, the inputs are visual tokens $\mathbf{X}\in \mathbb{R}^{ H \times W \times D}$ and attention mask $\mathbf{M_A} \in  \mathbb{R}^{ J \times J}$. The attention mask $\mathbf{M_A} \in  \mathbb{R}^{ HW \times HW}$ is initially set to an all-one matrix, and then extracted from certain layers through attention pruning. In this case, the attention mask generated from the $L_3^{th}$ layer will be used in $L_4^{th}$ and $L_5^{th}$ layers, and will be dynamically pruned and updated in the $L_6^{th}$ layer. Later in the second Transformer encoder, the inputs for MMSA are keypoint tokens $\mathbf{X}_k \in \mathbb{R}^{J \times D}$ and the self-defined joint mask $\mathbf{M_J} \in  \mathbb{R}^{ J \times J}$. The calculation is the same as Equation\ref{eq:attention1}.

\subsection{Visual Token Attention Prune Module}
We introduce sparsity at the attention level on visual tokens by pruning attention connections. Specifically, for a given query token, we only calculate its attention connections with a small subset of selected tokens. The attention mask aims at determining the attention weights between different input tokens. Since attention maps are typically sparse, it is unnecessary to pursue intensive attention, and pruning it can selectively remove connections that are deemed less important.
%In our approach, we utilize the attention vector of each patch token to determine critical connections, without introducing any additional modules. 

Prior versions of ViT have relied on predetermined sparse patterns, such as neighborhood windows\cite{Lu2021} or dilated sliding windows\cite{Beltagy2020}, which can be limited for objects with varying shapes and sizes. Alternatively, some input-dependent methods have used deformable attention to find corresponding tokens\cite{Zhu2020}, but this requires additional modules to learn the selected tokens. The Tri-level ViT\cite{Kong2022TriLevel} proposed a relatively feasible scheme, but no concrete experiments have been carried out on it. This paper intends to implement the method of human posture detection and explore its influence on model accuracy and efficiency. We employ the MMSA mechanism in the transformer to better capture the complexity of images, as shown in Fig.\ref{fig:framework1}.

\begin{figure}[h]
    \centering
    \includegraphics[width=0.9\linewidth]{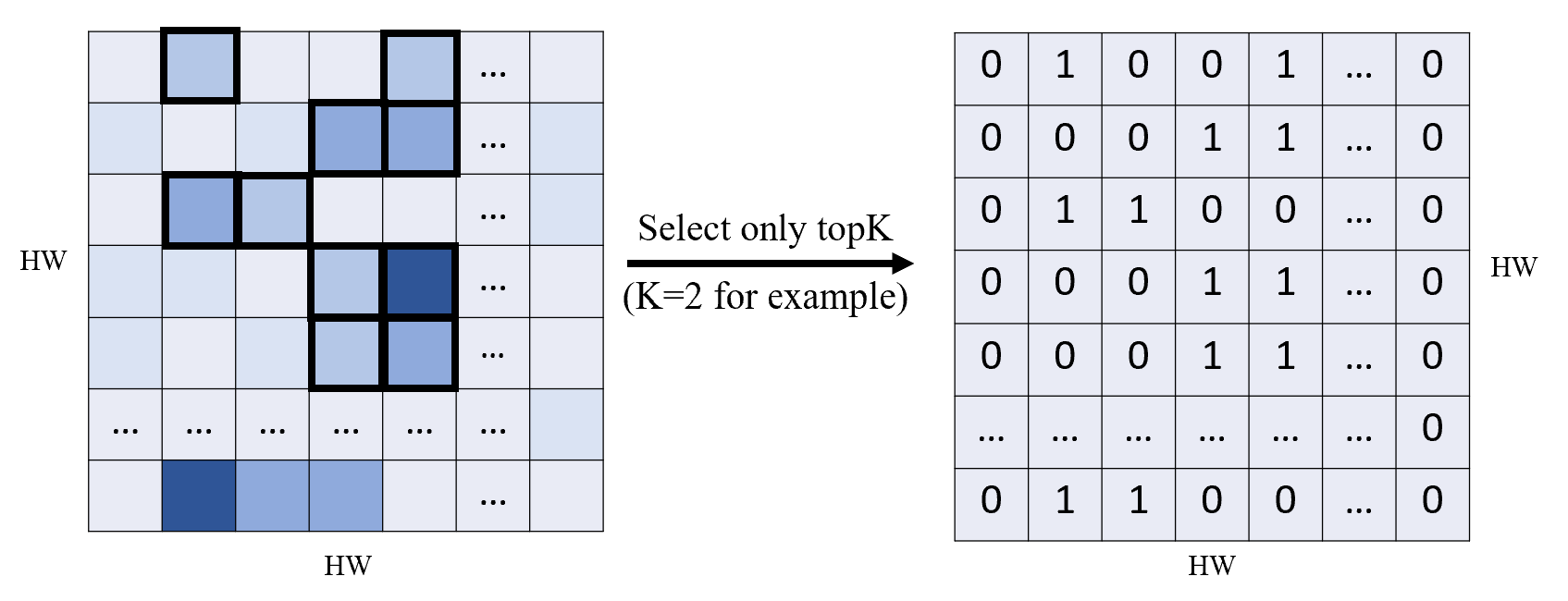}
    \caption{\small{The extraction of sparse attention mask }  }
    \label{fig:attn_map}
\end{figure}

As shown in Fig.\ref{fig:attn_map}, given the image matrix $\mathbf{X}$ and $\mathbf{M}$, for each row we dynamically select the top K values and form a new 0-1 attention mask. After the calculation and normalization of the self-attention matrix, we can simplify the attention mask. By dynamically choosing necessary tokens and pruning the less relevant smaller numbers, we can remove the redundant data to improve efficiency.

We initialize all elements in the mask to 1 and update the mask progressively. The prediction modules take the current $\mathbf{M}\in \mathbb{R}^{N \times N}$ (When considering joint mask as input the mask size is $\mathbf{M}\in \mathbb{R}^{N \times C}$) and the tokens $\mathbf{X}\in \mathbb{R}^{N\times D}$ as input. $\mathbf{N}$ stands for $\mathbb{H \times W}$. We compute the weighted sum over all values $v$ in the sequence tokens, together with their respective query $q^i$ and key $k^j$ representations.

%qkv
\begin{equation}
  \mathbf{[q,k,v] = zU_{qkv},\quad U_{qkv}\in \mathbb{R}^{D\times3D_h}}
  \label{eq:qkv}
\end{equation}

The attention weights $A_{ij}$ are based on not only the pairwise similarity between two elements of the sequence and their respective $q^i$ and $k^j$, but also the Attention mask to progressively update the more important attention weights. We will multiply $A$ with our input map $M$ to get a simpler matrix $A'$

%Aij
\begin{equation}
    \mathbf{A'=\mathbf{A}\times\mathbf{M},\quad A\in \mathbb{R}^{N\times N}, \;M\in \mathbb{R}^{N \times N}}
    \label{eq:attention}
\end{equation}

Notice that we are only doing softmax when the value in the matrix $M_{ij}$ is 1. If $M_{ij}$ is 0 we will skip the calculation process, making the operation easier.
%softmax
\begin{equation}
    \text{softmax}\mathbf{(A'_{ij})=\frac{exp(A'_{ij})}{ {\textstyle \sum_{j=0}^{J}exp(A'_{ij})(\text{only \, when} \; M_{ij}=1)} } }
    \label{eq:softmax}
\end{equation}

The superiority of the attention prune module is manifested in two aspects. Firstly the attention mask is generally very sparse, there is no need to consider all of them. Secondly, it is approved that attention is not the bigger the better. Some researchers tended to throw away all the attention, but this may result in losing information about the image. So in our method, we are trying to make it shorter while still keeping the valid information, in order to achieve substantial performance improvement. According to the analysis of each operation in a Transformer block \cite{kong2022spvit}, the pruning operations are performed on weight tensors of each attention head in the Multi-Self Attention (MSA) module. Given an input sequence, $N \times D$, in which $N$ is the input sequence length or the token number, and $D$ is the embedding dimension of each token. In MSA, the input size was originally $N \times N$, but by doing visual token attention pruning, we can reduce the input size to $N \times D$, further simplifying the calculation.

\subsection{Joint Attention Mask}
\begin{figure}[h]
    \centering
    \includegraphics[width=0.8\linewidth]{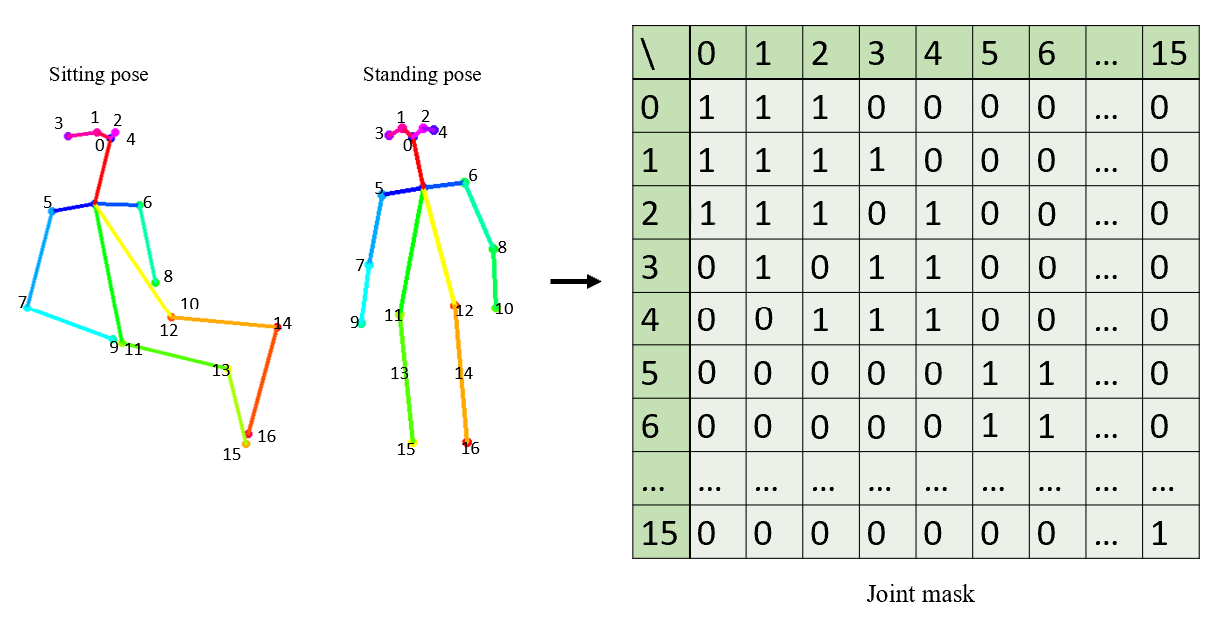}
    \caption{\small{The extraction of human joint mask }  }
    \label{fig:jointmask}
\end{figure}

As shown in Fig.\ref{fig:jointmask}. No matter what the posture of the human body is, the connection between the joints is always there and has a certain pattern. Therefore we can be able to summarize and pre-define a keypoint joint mask $\mathbf{M_J} \in  \mathbb{R}^{ J \times J}$ based on the human graph to emphasize the connection between limbs and strengthen the joint relationship during training. By using the sparsity of the junction, for each human joint, we only keep its adjacent nodes and symmetric nodes, set them to 1, and set the other nodes to 0, to form a sparse joint mask. For instance, for joint number $5$, the left shoulder, we only keep its neighboring left elbow joint and its symmetric joint, the right shoulder.

A corresponding operation is performed on each joint of the human body structure to form a  ${J \times J}$ mask. It should be used as a constant value and add to the whole process of training. The joint mask will later fed into the Graph Transformer layer(as in Fig.\ref{fig:framework1}) to emphasize the invariability of human joint connection.

\subsection{Implementation Detail}
On the Improved TokenPose with Sparsity, the number of encoder layers for Transformer $L$ is set to $12$, the embedding size $D$ is set to $192$, and the number of heads $H$ is set to $8$. And in Graph Transformer the number of encoder layers $L'$ is set to $8$. They take the shallow stem-net and the HRNet-W32 as the CNN backbone.

\section{Experiments and Validation}
\subsection{Settings} 
We evaluate our Improved TokenPose with Sparsity on monocular 2D human pose estimation benchmarks, datasets MPII\cite{andriluka14cvpr} 
%and COCO\cite{lin2014microsoft}. COCO comprises a vast collection of $200K$ images across $80$ object categories, with over $250K$ person instances annotated for keypoint detection tasks and its annotations encompass the locations of $17$ keypoints. MPII
contains about $25K$ images and $40K$ human instances with $16$ keypoints. The evaluation is based on the head-normalized probability of correct keypoint (PCKh) score \cite{andriluka14cvpr}. A keypoint is correct if it falls within a predefined threshold to the ground-truth location. We report the mean PCKh score by convention.

\subsection{Ablation Studies}

\begin{table}[t]
\centering
\resizebox{1.0\textwidth}{!}{
\begin{tabular}{|l|c|c|c|c|c|c|c|c|c|}
\toprule
\textbf{Method} & Head & Sho & Elb & Wri & Hip & Kne & Ank & Mean & Mean@0.1  \\
\hline\hline
TokenPose-S \cite{li2021tokenpose}  & 96.3 & 93.9 & 86.1 & 79.2 & 86.2 & 79.6 & 74.5 & 85.8 & 30.9\\
\hline
TokenPose with joint mask &96.3 & 94.2 & 87.1 & 80.0 & 85.8 & 80.8 & 75.8 & 86.3 & 31.17  \\
\hline
Improved TokenPose( AKR = 0.2 ) & 95.8 & 93.3 & 85.4 & 78.3 & 85.4 & 79.4 & 74.2 & 85.2 & 29.98 \\

Improved TokenPose( AKR = 0.5 ) & 96.1 & 94.3 & 87.3 & 80.3 & 86.7 & 81.1 & 76.2 & 86.7 & 31.23 \\

Improved TokenPose( AKR = 0.6 )  & 96.1 & 94.0 & 87.0 & 80.6 & 86.9 & 81.4 & 75.9 & \textbf{86.8} & \textbf{31.28} \\

Improved TokenPose( AKR = 0.7 )  & 96.0 & 94.0 & 86.7 & 79.8 & 85.8 & 80.0 & 73.9 & 85.9 & 31.24 \\

Improved TokenPose( AKR = 0.8 ) & 96.3 & 94.5 & 87.0 & 80.1 & 86.8 & 81.2 & 75.5 & 86.6 & 31.11 \\

\bottomrule
\end{tabular}}
\caption{ \small{Ablation studies on different Attention-Keep Ratio(AKR)}}
\label{tab:mpii_2d}
\end{table}

We first build our Improved TokenPose with Sparsity based on TokenPose-S for comparison. We follow the same training recipes as TokenPose\cite{li2021tokenpose}. Table \ref{tab:mpii_2d} shows the progressive results of the human pose evaluation. Adding the joint mask can bring a slight improvement. Making use of the sparsity in both visual token attention and keypoint token attention can have an improvement of at least one percentage point. As shown in the results, the accuracy is the best when the attention-keep ratio is 0.6, which means focusing only 60 percent of the visual tokens and throwing away the rest of them.  

The results in Table \ref{tab:mpii_2d} proved that the pruned visual token connections might correspond to less critical relationships, and help with highlighting the more influential relationships in the human pose estimation process. However, it's important to note that the pruning process may impact the overall accuracy of the pose estimation model. There is definitely a trade-off. Removing certain connections may discard valuable information, making the model lose some fine-grained details of human poses. It still requires careful analysis and experimentation to strike a balance between pruning for efficiency and preserving accuracy.

\subsection{Results}

\begin{table}[h!]
% >>>>>>>>>>>>>>>>>>>>>>>>>>> MPII >>>>>>>>>>>>>>>>>>>>>>>>>>>
\centering
\resizebox{1.0\textwidth}{!}{
\begin{tabular}{|l|lllllll|c|}
\toprule
Method  & Head & Sho & Elb & Wri & Hip & Kne & Ank & Mean  \\
\hline\hline
SimpleBaseline-Res50 \cite{xiao2018simple}  & 96.4 & 95.3 & 89.0 & 83.2 & 88.4 & 84.0 & 79.6 & 88.5  \\
HRNet-W32. \cite{sun2019deep}               & 96.9 & 96.0 & 90.6 & 85.8 & 88.7 & 86.6 & 82.6 & 90.1 \\
\hline
TokenPose-S \cite{li2021tokenpose}  & 96.3 & 93.9 & 86.1 & 79.2 & 86.2 & 79.6 & 74.5 & 85.8 \\
Improved TokenPose-S                                  &96.1 & 94.1 & 87.0 & 80.6 & 87.0 & 81.4 & 75.9 & 86.8(+\textbf{1.0})\\

\bottomrule
\end{tabular}
}
\caption{\small{Results on the MPII validation set. The input size is $256\times256$.}}
\label{tab:mpii}
\end{table}

The experimental results on the MPII dataset are shown in Table \ref{tab:mpii}. As we can see our methods have some improvements compared with the original TokenPose baseline. The idea of making use of attention-level visual tokens and keypoint tokens sparsity is proven to be efficient and effective, and this idea can be applied to other fields as well.

\begin{figure}[ht]
    \centering
    \includegraphics[width=0.8\linewidth]{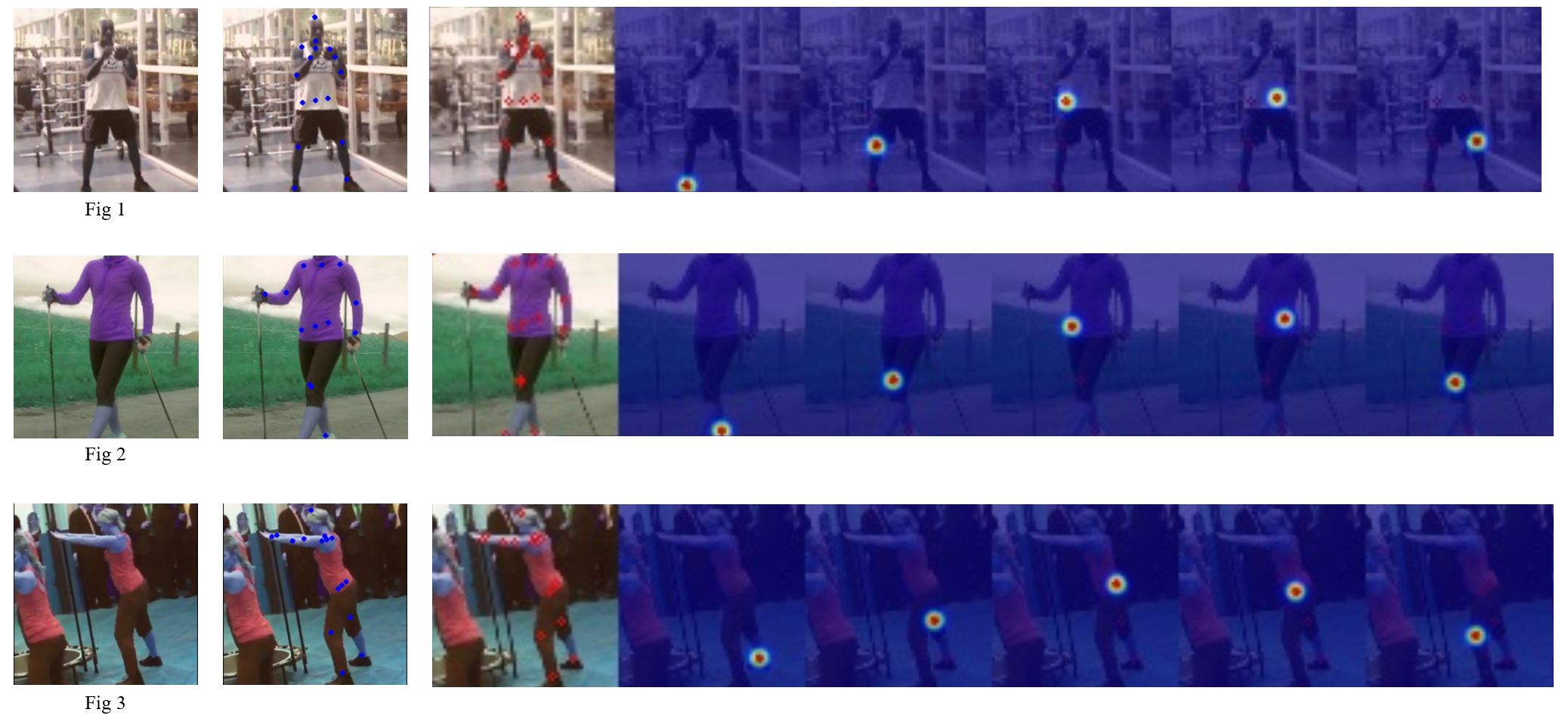}
    \caption{\small{Sample heatmaps of our approach}  }
    \label{fig:result1}
\end{figure}

\subsection{Visualizations}
\begin{figure}[h!]
    \centering
    \includegraphics[width=0.7\linewidth]{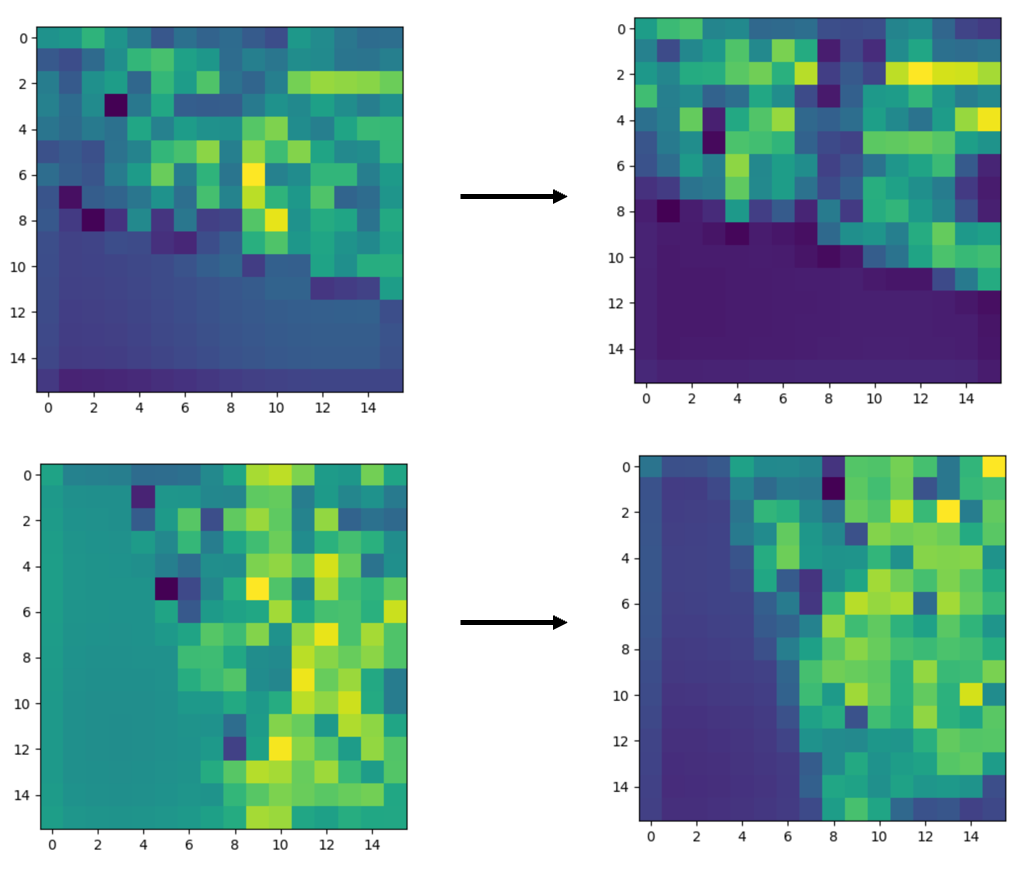}
    \caption{\small{The process of attention pruning}  }
    \label{fig:prune}
\end{figure}

\begin{figure*}[ht]
    \centering
    \includegraphics[width=1.\linewidth]{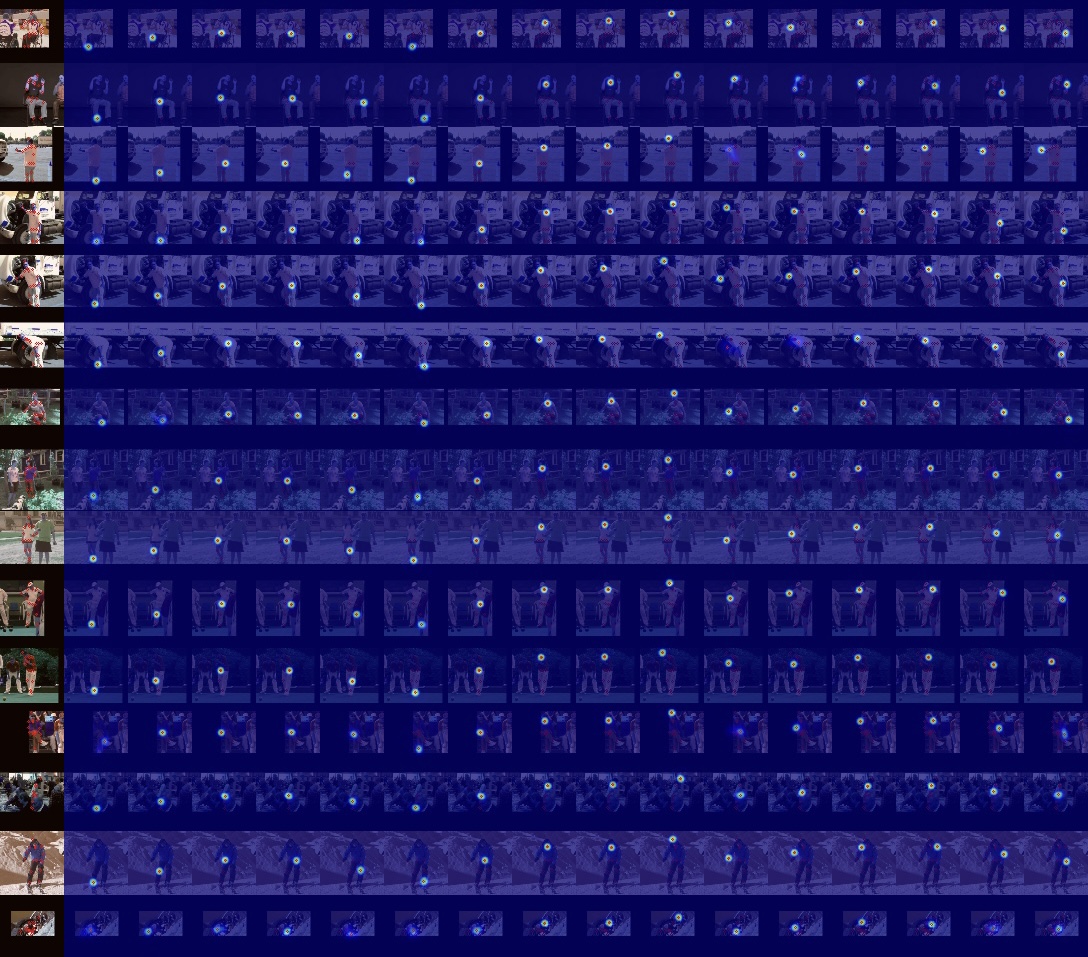}
    \caption{\small{Heatmaps with recognized joints on MPII dataset}  }
    \label{fig: extra}
\end{figure*}

We visualize the sample heatmaps of our approach in Fig.\ref{fig:result1} and Fig.\ref{fig: extra}. We visualize the procedure of attention pruning. Fig.\ref{fig:prune} shows how the attention maps are pruned are dynamically select the important information during the procedure. After the pruning, we will drop a few tokens that don't make much difference to the final result. 

\newpage
\section{Conclusion}
In this paper, we propose the Improved TokenPose with Sparsity for 2D human pose estimation. Our superiority is reflected in two parts: the combination of human body structure, and the use of attention pruning to generate a simpler attention map. We reached improvements for applying the sparsity in keypoint and visual tokens. Experiments on the MPII dataset show that the Improved TokenPose with Sparsity achieves better accuracy compared with previous TokenPose networks. 

%Bibliography
\newpage
\bibliographystyle{unsrt}  
\bibliography{references}  

\end{document}